%% file: bare_jrnl.tex
\documentclass[journal]{IEEEtran}
\usepackage{times}
\usepackage{epsfig}
\usepackage{graphicx}
\usepackage{amsmath}
\usepackage{amssymb}
\input{math_commands.tex}

\ifCLASSINFOpdf
  \DeclareGraphicsExtensions{.pdf,.jpeg,.png}
\else
\fi
\usepackage{bm}
\usepackage{svg}
\usepackage{amsmath}
\usepackage{amssymb}
\usepackage{algorithm}
\usepackage{algpseudocode}
\usepackage{placeins}
\usepackage{float}
\usepackage{hyperref}
\usepackage{soul}
\usepackage{booktabs}
\usepackage{diagbox}

\algnewcommand\algorithmicinput{\textbf{Input:}}
\algnewcommand\INPUT{\item[\algorithmicinput]}

\begin{document}
%
% paper title
% Titles are generally capitalized except for words such as a, an, and, as,
% at, but, by, for, in, nor, of, on, or, the, to and up, which are usually
% not capitalized unless they are the first or last word of the title.
% Linebreaks \\ can be used within to get better formatting as desired.
% Do not put math or special symbols in the title.
\title{Effect of The Latent Structure on Clustering with GANs}
%
%
% author names and IEEE memberships
% note positions of commas and nonbreaking spaces ( ~ ) LaTeX will not break
% a structure at a ~ so this keeps an author's name from being broken across
% two lines.
% use \thanks{} to gain access to the first footnote area
% a separate \thanks must be used for each paragraph as LaTeX2e's \thanks
% was not built to handle multiple paragraphs
%

\author{Deepak Mishra, Aravind Jayendran and Prathosh A. P. \thanks{Deepak is with IIT Jodhpur, Aravind is with Flipkart Internet Pvt Ltd and Prathosh is with IIT Delhi, India. (e-mail: dmishra@iitj.ac.in, aravind.j@flipkart.com, and prathoshap@gmail.com.)}}

% note the % following the last \IEEEmembership and also \thanks - 
% these prevent an unwanted space from occurring between the last author name
% and the end of the author line. i.e., if you had this:
% 
% \author{....lastname \thanks{...} \thanks{...} }
%                     ^------------^------------^----Do not want these spaces!
%
% a space would be appended to the last name and could cause every name on that
% line to be shifted left slightly. This is one of those "LaTeX things". For
% instance, "\textbf{A} \textbf{B}" will typeset as "A B" not "AB". To get
% "AB" then you have to do: "\textbf{A}\textbf{B}"
% \thanks is no different in this regard, so shield the last } of each \thanks
% that ends a line with a % and do not let a space in before the next \thanks.
% Spaces after \IEEEmembership other than the last one are OK (and needed) as
% you are supposed to have spaces between the names. For what it is worth,
% this is a minor point as most people would not even notice if the said evil
% space somehow managed to creep in.

% The paper headers
\markboth{}%
{Shell \MakeLowercase{\textit{et al.}}: Bare Demo of IEEEtran.cls for IEEE Journals}
% The only time the second header will appear is for the odd numbered pages
% after the title page when using the twoside option.
% 
% *** Note that you probably will NOT want to include the author's ***
% *** name in the headers of peer review papers.                   ***
% You can use \ifCLASSOPTIONpeerreview for conditional compilation here if
% you desire.

% If you want to put a publisher's ID mark on the page you can do it like
% this:
%\IEEEpubid{0000--0000/00\$00.00~\copyright~2015 IEEE}
% Remember, if you use this you must call \IEEEpubidadjcol in the second
% column for its text to clear the IEEEpubid mark.

% use for special paper notices
%\IEEEspecialpapernotice{(Invited Paper)}

% make the title area
\maketitle

\maketitle

\begin{abstract}
Generative adversarial networks (GANs) have shown remarkable success in generation of data from natural data manifolds such as images. In several scenarios, it is desirable that generated data is well-clustered, especially when there is severe class imbalance.
In this paper, we focus on the problem of clustering in generated space of GANs and uncover its relationship with the characteristics of the latent space. We derive from first principles, the necessary and sufficient conditions needed to achieve faithful clustering in the GAN framework: (i) presence of a  multimodal latent space with adjustable priors, (ii) existence of a latent space inversion mechanism and (iii) imposition of the desired cluster priors on the latent space. We also identify the GAN models in the literature that partially satisfy these conditions and demonstrate the importance of all the components required, through ablative studies on multiple real world image datasets.  Additionally, we describe a procedure to construct a multimodal latent space which facilitates learning of cluster priors with sparse supervision. The code for the implementation can be found at \textit{https://github.com/NEMGAN/NEMGAN-P}
\end{abstract}

\section{Introduction}
\subsection{Background and Contributions}
Generative Adversarial Networks(GANs)~\cite{goodfellow2014generative,arjovsky2017wasserstein,miyato2018spectral,brock2018large,mao2017least} and its variants are a category of highly successful generative neural models which learn mappings from arbitrary latent distributions to highly complex real-world distributions. In several downstream tasks such as conditional generation, data augmentation and class balancing ~\cite{hastie2009unsupervised,bengio2013representation,radford2015unsupervised,kingma2014semi,cheung2014discovering}, it is desirable that the data generated by a generative model is clustered. However, it is well known that GANs in their raw formulation are unable to fully impose all the cluster properties of the real-data on to the generated data
\cite{arora2017gans,wu2017gp,karras2017progressive,brock2018large}, especially when the real-data has skewed clusters. %Motivated by these observations, in this work, we intend to formulate and address the following question - \textbf{\textit{How to achieve clustering along with good quality data generation in a GAN when there is class-imbalance with unknown class priors?}}
While a lot of efforts have been devoted in past to stabilize the GAN training~\cite{salimans2016improved,gulrajani2017improved,karras2017progressive}, little attention has been given to understand the impact of latent space characteristics on data generation (a brief review of related methods is given in Sec. 4). Motivated by these observations, we propose to accomplish the following:

\begin{enumerate}

    \item {Starting from the first principles, formulate the necessary and sufficient conditions needed for faithful clustering in the generated space of GAN.}
	\item {Demonstrate the importance of each of the condition through ablative studies using different GAN models that partially satisfy them, on four large-scale datasets. }
	\item {Propose a method for the construction of a learnable multi-modal latent space that facilitates sparsely supervised learning of the cluster priors.}
	
\end{enumerate}

\subsection{Problem setting}
%Figure.~\ref{bd} shows overview of the proposed method. It presents a model for GANs where an inversion mapping from the generated data space to an engineered latent space is learned such that properties of the data generating distribution are matched to those of the latent distribution.
In the context of GANs, clustering in generated space refers to inheritance of the cluster properties of real data on the generated data. In GANs, a generator $g$, which is a function of the latent variable ($\rvz$), is tasked to sample from the desired data ($\rvx$) distribution via an adversarial game \cite{goodfellow2014generative}. Suppose $P_{X}$  and $P_{W}$, respectively be the distributions of the generated and real data, with  $\sX$ and  $\sW$ representing their support. We call a distribution clustered (or equivalently multi-modal) if its support is a disconnected union of non-empty pairwise-disjoint connected open subsets. For instance, $P_{W}$ is a clustered distribution with $M$ clusters if $\sW$ is a disconnected union of $M$ non-empty pairwise disjoint connected open subsets $\Big(\sW_i, i\in\{0,1,...,M-1\}\Big)$ and $\sW \equiv \bigcup_{i=0}^{M-1}{\sW}_i$, with $\sW_i$ denoting the support of the $i^{th}$ mode \footnote{For simplicity, we have assumed that the clusters do not overlap albeit all the analysis can be extended to the case where clusters have minimal overlap.}. With this definition, clustering in generated space amounts to the following: if $P_{W}$ is clustered in the aforementioned way, $P_{X}$ is also clustered in the exact same way. That is, the probability masses of $P_{W}$ and $P_{X}$ over each individual cluster (or mode) are the same.

%Given a clustered real data,, with corresponding distribution , we desire that $P_{X}$ is a multimodal distribution having modal properties same as $P_{W}$, where each mode corresponds to a cluster. Mathematically, the support ${\sX}$ of $P_{X}$ is a disconnected set. It is a union of $M$ non-empty pairwise disjoint connected open subsets $(\sX_i, i\in\{0,1,...,M-1\})$, when there are $M$ modes with $\sX_i$ denoting the support of the $i^{th}$ mode. Further, $\sW \equiv \bigcup_i^{M-1}{\sW}_i$, the support of $P_{W}$, is also a disconnected set and 

\section{Clustering in GANs - Requirements}
Firstly, we show that to obtain a clustered generated space or equivalently, a multimodal $P_{X}$, it is necessary to have a multimodal latent space $ P_{Z} $ with a structure similar to the real-data.

\textbf{Lemma 1:} Let $\sZ$ denote the support of $ P_{Z} $. If $ \sZ_i\subseteq \sZ$ denote the inverse images of ${\sX}_i$ under $g$, then $ \bigcap_i{\sX}_i = \Phi$ only if $ \bigcap_i{\sZ}_i = \Phi$, where $\Phi$ is an empty set. 

\textbf{Proof:} Without the loss of generality we assume $M = 2$. Assume $\displaystyle \sX_0 \cap \sX_1 = \Phi$ and  $\displaystyle \sZ_0 \cap \sZ_1 \neq \Phi \implies  \exists \rvz_i \in \sZ_0 \cap \sZ_1$.
Given $\displaystyle \rvz_i \in \sZ_0$, let $g(\rvz_i) = \rvx_{i0} \in \sX_0$ and similarly, given $\displaystyle \rvz_i \in \sZ_1$, $g(\rvz_i ) = {\rvx}_{i1} \in {\sX}_1$. Since $g$ is a continuous function, ${\rvx}_{i0} = {\rvx}_{i1} = {\rvx}_{i} \implies {\rvx}_{i} \in \sX_0 \cap \sX_1$ contradicting the fact that $\sX_0 \cap \sX_1 = \Phi$, hence $\displaystyle \sZ_0 \cap \sZ_1 = \Phi.$ 

%\break \blacksquare$

Even though a multimodal latent space is a necessary condition (Lemma 1), it is not sufficient. The generating function $g$ can be non-injective, implying that multiple modes of the latent space can collapse to a single mode in the generated space. However, if there exists another continuous mapping $ h:{\sX}\rightarrow {\hat{\sY}}$ which maps the generated samples $\rvx$, to another auxiliary random variable $\mathbf{\hat{y}}$ such that $P_{\hat{Y}}$ is also multi-modal, then Lemma 1 can be applied again on $h$ to guarantee multimodality on $P_{X}$, as stated in the following corollary to Lemma 1.

\textbf{Corollary 1.1. }
Let  $ h:\sX\rightarrow \hat{\sY}$ and $\hat{\sY}_i\subseteq \hat{\sY}$ be a subset of $\hat{\sY}$. Then $ \bigcap_i\hat{\sY}_i  = \Phi$ only if $ \bigcap_i{\sX}_i  = \Phi$.  Given $\bigcap_i{\sZ}_i = \Phi$, the condition $\bigcap_i\hat{\sY}_i = \Phi$ is sufficient for $ \bigcap_i{\sX}_i= \Phi$.

Corollary 1.1 states that if the latent distribution ($ P_{Z} $) is multimodal with $M$ modes and $h$ maps $\rvx$ to any multimodal distribution ($P_{\hat{Y}}$) with $M$ modes, the generated distribution, $P_X$, will also have $M$ modes. 
Even though in principle, it is sufficient that if $P_{\hat{Y}}$ is any $M$ modal distribution to achieve clustering in $P_X$, the clusters may not be optimal as ascertained in the following corollary. 

\textbf{Corollary 1.2.} Suppose $g$ is the generator network of a GAN which maps $P_Z$ to $P_X$ and $h$ is an inverter network which maps $P_X$ to $P_{\hat{Y}}$. Further, let us assume all the distributions, $P_Z$, $P_X$ and $P_{\hat{Y}}$, along with the real data distribution $P_W$ are multimodal with $M$ modes having disjoint supports. The cluster properties of the real data $\sW$ will not be reflected in the generated data $\sX$, if the probability mass under every mode (cluster) in $P_Z$ does not match with the modal masses of $P_W$ (Proof in the Supplementary material).

%\begin{align}
%    \int_{{\sW}_i}P_W~d\rvw &\neq \int_{{\sZ}_i}P_Z~d\rvz\\
% \implies \int_{{\sW}_i}P_W~d\rvw &\neq \int_{{\sX}_i}P_X~d\rvx
%\end{align}
Thus, if either $P_Z$ or $P_{\hat{Y}}$ are chosen such that their mode (cluster) masses do not match with that of real data distribution $P_W$, the adversarial game played in the GAN objective cannot force $P_X$ to follow $P_W$. In other words, cluster properties of the real data $\sW$ will not be reflected in the generated data $\sX$ leading to incorrect coverage of the clusters in the generated data as observed in \cite{khayatkhoei2018disconnected}.
In summary, the following are the necessary and sufficient conditions required to accomplish clustering in the generated space of a GAN.
\begin{enumerate}

    \item The latent space which is the input to the GAN, should be multimodal with number of modes equal to the number of clusters in the real data ($\mathbf{C_1}$).
    \item There should be a continuous mapping from the generated space to an auxiliary multimodal random variable with same cluster properties as the real data ($\mathbf{C_2}$).
    \item The mode (cluster) masses of the distributions of the latent and auxiliary variables must match to the mode masses of the distribution of the real data ($\mathbf{C_3}$).   
\end{enumerate}
%Motivated by these observations, we describe a generic method to construct a multimodal latent space with adjustable mode masses and a latent inversion procedure to achieve faithful clustering in the generated space of the GAN. 

%Note that condition 3, requires that the class priors need to be learned, when it is not known. Figure 1 and Table 1 illustrate the facts mentioned in this Sectionby learning different models of GANs that violate at least one of the three conditions, on a two-class imbalanced toy dataset (interspersed Moons). It can be seen that the GANs that violate one of the three conditions fails to cluster the generated data as corroborated with the cluster purity measures. 

\section{Clustering in GANs - Realization}

%Thus, we introduce a new random variable $\rvy$ that is an indicator of the modes of $\rvz$. Formally,  $\rvy:= \1_\mathrm{z\in\sZ_1}$,  which implies that $P_{Y}(\rvy = 1 ) = \int_{ \sZ_1 }P_{Z} ~d\rvz$ and $P_{Y}(\rvy = 0) = \int_{\sZ_0}P_{Z}~d\rvz$. 
%The aforementioned lemma and corollary ensures multimodality in $P_X$ albeit it doesn't make the mode mass of $P_X$ and $P_Z$ the same.  
%In this section, we describe procedures to realize all the conditions mentioned in the previous section. The idea is to construct a multimodal latent space with learnable priors that would latch on to the real class priors post learning. Further, a mapping from the generated space is learned on to an auxiliary distribution that serves as a cluster indicator while also aiding class prior learning with sparse supervision.  

In this section, we describe the possible methods for realizing the requirements for clustering with GANs.

\subsection{Multimodal Latent space}

%Engineering the latent space in accordance with the true data distribution is the first problem of interest. In this work, we target matching two modal properties namely  1) the number of modes and 2) modal mass or mode priors. We desire the latent distribution to have same number of modes with their corresponding priors as the true data distribution. Let the latent space be represented by $\sZ$ and $P_{Z}$ denote its distribution. We define $P_{Z}$ as a multimodal distribution with $M$ modes, if its support, ${\sZ}$ is a disconnected set, which is a union of $M$ non-empty pairwise disjoint connected open subsets $(\sZ_i, i\in\{0,1,...,M-1\})$, each representing a mode.  
Two known ways of constructing a multimodal latent space are 1) using the mixture of continuous distributions such as GMM~\cite{gurumurthy2017deligan}, 2) using the mixture of a discrete and a continuous distribution~\cite{chen2016infogan, mukherjee2018clustergan}. Latter one is more popular and often realized by concatenation of discrete and continuous random variables. We describe a more general form of this by using an additive mixture of a pair of discrete and continuous random variables, which facilitates flexible mode priors. 

Let the latent space be represented by $\sZ$ and $P_{Z}$ denote its distribution with $M$ modes.
This could be obtained by taking an additive mixture of a generalized discrete distribution and a compact-support continuous distribution such as uniform distribution. Let $\rvy \sim P_{Y}$  and $\boldsymbol{\nu_2} \sim P_{N_2}$ denote samples drawn from the discrete and continuous distributions, respectively. Accordingly, the latent space $\rvz$ is obtained as: $\rvz =  \rvy + \boldsymbol{\nu_2}$. This results in a multi-modal continuous distribution with disconnected modes since $P_Z = P_Y * P_{N_2}$, where $*$ denotes the convolution product. The support of $P_{N_2}$ is chosen in such a way that the modes of $P_Z$ are disjoint.  In $P_Z$, the number and the mass of the modes are obtained from discrete component $(P_Y)$ and the continuous component $(P_{N_2})$ ensures variability.  The discrete component $\rvy\sim P_Y$ can also be interpreted as an indicator of the modes of $\rvz$. Formally,  $\rvy:= i \quad\forall {z\in\sZ_i}$,  which implies that $\int_{ \sZ_i }P_{Z} ~d\rvz = P_{Y}(\rvy = i ) \label{mode}$.

Note that in all the aforementioned latent space construction strategies, the latent space parameters are fixed and cannot be changed or learned to suit the real-data distribution. To alleviate this problem, we propose to reparameterize a second continuous uniform distribution, $P_{N_1}$, using a vector $\boldsymbol{\alpha}$ to construct the desired $P_Y$.
Let $\boldsymbol{\alpha} = [\alpha_0,\alpha_1,.....,\alpha_{M-1}]^T$, $\alpha_i \in \mathbb{R}$ be an arbitrary vector and   $\nu_1 \sim P_{N_1}(\nu_1) = \mathbb{U}[0,1] $. We define a function, $f\left(\boldsymbol{\alpha}, \nu_1\right): \sR^M\times \sR \rightarrow \sR^M$ reparameterizing  $P_Y$ as follows.
\begin{equation}
    f_i\left(\alpha_i, \nu_1\right) = \begin{cases}\sigma_h\left( a_i-\nu_1\right) - \sigma_h\left(a_{i-1}-\nu_1\right); &i\neq 0\\
    \sigma_h\left( a_i-\nu_1\right);  &i=0
    \end{cases}
    \label{f1}
\end{equation}
where$f_i$ is the $i^{th}$ element of $f$, $\sigma_h$ is a unit step  function and $a_i$ is given as
\begin{equation}
    a_i = \frac{1}{{\sum_{k}{e^{\alpha_k}}}}\sum_{j = 0}^{i}{e^{\alpha_i}}
\end{equation}

With these, one can reparametrize a uniform distribution using $\boldsymbol{\alpha}$ and $f$, to obtain a multinoulli distribution. 

\textbf{Lemma 2:} Define $\rvy =: \argmax_{i\in \{0,..,M-1\}} f_{i}$, then $\rvy$ follows a multinoulli distribution $P_Y$  with
\begin{equation}
    P_Y(\rvy=i) = \frac{e^{\alpha_i}}{\sum_{k}{e^{\alpha_k}}}
    \nonumber
\end{equation} (Proof in the supplementary material).

% \textbf{Proof:} Since $\sigma_h$ is a unit step function, $f$ is the first order difference or discrete Dirac delta function positioned at $a_i$.
% %\begin{equation}
% %    f_{1_i} = \delta(a_i - \nu_1)
% %    \label{delta}
% %\end{equation}
% Now by definition,
% \begin{equation}
%     P_Y(\rvy=i) = P(f_{i} \neq 0)
%     \label{prior}
% \end{equation}
% From \eqref{f1}, we can see that $f_{i}$ becomes non-zero only for $a_{i-1}\leq\nu_1\leq a_i$, therefore,

% \begin{align}
% P_Y(\rvy=i) &= P_{N_1}(a_{i-1}\leq\nu_1\leq a_i)\\
%      &= \int_{a_{i-1}}^{a_i}P_{N_1}(\nu)d\nu = a_i - a_{i-1} = \frac{e^{\alpha_i}}{\sum_{k}{e^{\alpha_k}}}\label{alpha}
%   \end{align}

Therefore, starting from an arbitrary discrete valued real vector and sampling from a known uniform distribution, one can obtain a multinoulli random variable whose parameters become a function of the chosen arbitrary discrete vector $\boldsymbol{\alpha}$ which may be fixed according to the real data or learned through some inductive bias.%From an implementation perspective, we approximate $\sigma_h$ with a hard sigmoid to enable backpropogation. 

% \begin{figure}[!t]
% \centering
% {\includegraphics[width=8cm]{BD.png}}
% \caption{Illustration of the proposed method. Latent vector $\rvz$ is sampled from an additive mixture of a continuous $P_{N_2}(\boldsymbol{\nu_2})$ and discrete $P_Y(\rvy)$ distribution, resulting in a multimodal distribution $P_{Z}$. Generator $g(\rvz)$ tries to mimic the real data distribution $P_{X}$ with the help of discriminator $d$. The inversion network $h(g(\rvz))$ inverts the generation process to ensure the matching of clustering properties of generating and latent distributions. Mode priors of the latent space is encoded in $\rvy$ by reparameterizing a known distribution $P_{N_1}({\nu_1})$ using a learnable vector $\boldsymbol{\alpha}$.}
% \label{bd}
% \end{figure}

\subsection{Latent inverter}

\subsubsection{Clustering}

As mentioned in the previous sections, it is necessary to have a mapping from the generated data space to an auxiliary random variable that would have same mode masses as the real data. This can be ensured by choosing $h({\rvx})=\hat{\rvy}$ (a neural network) that would minimize a divergence measure, $\mathcal{D} (P_{\hat{Y}}, P_{Y} )$, such as KL-divergence, between the distribution of its output $\hat{\rvy}$ and the distribution of the discrete part of the latent space (${\rvy}$). Learning an $h$ this way, would not only lead to clustered generation, but also ensures that the modal (cluster) properties of the latent space (and thus real data) is reflected in the generated space as described in the following lemma:

\textbf{Lemma 3:}  Let $\hat{\rvx}$ be a discrete random variable  that is an indicator of clusters (modes) of $P_X$. That is,  $P_{\hat{\sX}}(\hat{\rvx} =i) = \int_{{\sX}_i}P_X~d\rvx$. Then minimization of KL divergence,  $\KL (P_{\hat{Y}} ||P_{Y} )$, leads to minimization of $\KL (P_{\hat{Y}}  || P_{\hat{X}})$. (Proof given in the supplementary material).

Note that $h(\rvx)$ or $\hat{\rvy}$ acts like a posterior of the cluster assignment conditioned on the generated data. Therefore if the parameters of the input latent distribution ($\alpha_i$'s) are chosen in accordance with the modal properties of the real data distribution, generated data space will be well-clustered within a standard GAN training regime. If $g$ is the Generator of a GAN with $d$ denoting the usual discriminator network and $h$ is a neural network operating on the output of the generator to produce $\hat{\rvy}$, the objective function to be optimized for faithful clustering is given by:
\begin{equation}
    \min_{g, h} \max_d \mathcal{L}\left(g, h, d\right)
    \label{obj}
\end{equation}
\vspace{-0.5cm}
\begin{multline}
    \hspace{-0.3cm}\mathcal{L}\left(g, h, d\right) =  \E_{\rvw} [ \log d(\rvw) ] + \E_{\rvz} [ \log\left( 1 - d \circ g(\rvz) \right)] \\  + \mathcal{D} (P_{\hat{Y}}, P_{Y})
    \label{obj_l}
\end{multline}
where $\rvw$ represents samples from the real data distribution. For implementation, cross-entropy for the KL-term in \eqref{obj_l} is used, since the entropy of $P_{Y}$ is a constant for a given dataset. A block diagram representing the learning pipeline is given in the Supplementary material (Fig. 2).

\subsubsection{Learning cluster priors}

The presence of the inverter network provides an additional advantage. It helps in learning the true mode (cluster) priors in presence of a favourable inductive bias~\cite{locatello2018challenging}. In our formulation, information about the mode-priors is parameterized through the vector $\boldsymbol{\alpha}$. Let there be a set of few labelled real data samples, call it $\sW_s$, which provides the required inductive bias. As observed previously, the network $h(\rvx)$ is an estimator of the posterior of the cluster assignments given the data, $P(\hat{\rvy}|\rvx)$. Thus, marginalizing the output of $h(\rvx)$ over all $\rvx$ amounts to computing $\E_{\rvx}[h(\rvx)] $, which provides an estimate of $P_{\hat{Y}}$. Analogous to $\E_{\rvx}[h(\rvx)] $, the quantity $\frac{e^{\alpha_i}}{\sum_{k}{e^{\alpha_k}}}$ provides an estimate of $P_Y$ (Lemma 2). If the assumed $\boldsymbol{\alpha}$ is incorrect, then $h$ would mis-assign cluster labels on some of $\sW_s$. In other words, $P_Y$ and $P_{\hat{Y}}$ aren't the same which would be the same if the priors were correct. In this scenario, we propose to retrain $h(\rvx)$ on $\sW_s$ using a cross-entropy loss so that it assigns correct cluster labels on all of $\sW_s$. Subsequently, we re-estimate $\E_{\rvx}[h(\rvx)]$ for an arbitrary subset of unlabelled data (typically less than 1\%), with the new $h(\rvx)$. Now since $h(\rvx)$ is changed (via retraining), one can use the mismatch between $\E_{\rvx}[h(\rvx)]$ and $\frac{e^{\alpha_i}}{\sum_{k}{e^{\alpha_k}}}$ to re-compute $\boldsymbol{\alpha}$. 

The following is the loss function used that incorporates the aforementioned idea for learning $\boldsymbol{\alpha}$:
\begin{equation}
    \min_{h, \boldsymbol{\alpha}}\mathcal{L}_\alpha = \min_{h}\mathcal{L}_{cc} + \min_{\boldsymbol{\alpha}}\big|\big|\E_{\rvx}[h(\rvx)] - \frac{e^{\alpha_i}}{\sum_{k}{e^{\alpha_k}}} \big|\big|_1
\end{equation}
%Accordingly the objective function is (\eqref{obj}) is updated as
%\begin{equation}
%    \min_{g, h} \max_d \mathcal{L}\left(g, h, d\right) + \min_{h}\mathcal{L}_{cc} + \min_{\boldsymbol{\alpha}}\mathcal{L}_\alpha
%\end{equation}
where $\mathcal{L}_{cc}$ is the cross-entropy loss used to train $h$ on $\sW_s$. \\ 
Note that prior-learning component is optional and independent of the GAN training which is completely unsupervised. However, since we have shown that with incorrect priors, GANs cannot cluster faithfully, the priors can be first learned, if unknown, and GANs can be trained with the correct priors.

\section{GAN Models}
\label{related}
In this section, we identify the GAN models that satisfy at-least one of the three conditions required for clustering. Vanilla GANs such as DCGAN \cite{radford2015unsupervised}, WGAN \cite{arjovsky2017wasserstein}, SNGAN \cite{miyato2018spectral} etc. satisfy none of the three conditions. Models such as DeliGAN \cite{gurumurthy2017deligan}, GANMM \cite{yu2018mixture}, MADGAN \cite{ghosh2018multi} constructs a multimodal latent space using mixture models to avoid mode-collapse, nevertheless they neither have a latent inverter (C2) nor mode-matching (C3). Latent inverter network (with different choices for $d$) has been incorporated in the context of regularizing GAN training in many models such as VEEGAN \cite{srivastava2017veegan}, BiGAN \cite{donahue2016adversarial}, ALI \cite{dumoulin2016adversarially}, CATGAN \cite{springenberg2015unsupervised} etc. While all of these have latent inverter with different objectives, they lack multimodal latent space (C1) and  prior-matching (C3). InfoGAN \cite{chen2016infogan} and ClusterGAN \cite{mukherjee2018clustergan} have both multimodal latent space and latent inverter (with a mutual information maximization cost for $d$) but not the mode-matching (C3). 

In the subsequent sections, we consider a representative model from all categories to demonstrate the role of all the conditions via ablations.  In a model,  a satisfied and an unsatisfied condition is respectively denoted with $\bf{C_i}$ and  $\hat{C_i}$. For this study, we consider WGAN for $\hat{C}_1\hat{C}_2\hat{C}_3$, DeliGAN for $\mathbf{C_1}\hat{C_2}\hat{C}_3$, ALI/BiGAN for $\hat{C_1}\mathbf{C_2}\hat{C_3}$, InfoGAN/ClusterGAN for $\mathbf{C_1}\mathbf{C_2}\hat{C_3}$, and finally build a model (with WGAN as the base) with the described multimodal latent space, latent inverter (with KL-divergence for $h$) and matched prior for $\mathbf{C_1}\mathbf{C_2}\mathbf{C_3}$. For all the experiments, the class prior is fixed either to uniform (for $\hat{C_3}$) or matched to the appropriate mode/cluster prior (for $\mathbf{C_3}$), which provides the required inductive bias.  The underlying architecture and the training procedures are kept the same across all models. All GANs are trained using the architectures and procedures described in the respective papers.

%\subsubsection{Modified GAN learning for mode discovery}

%Salimans et. al. \cite{salimans2016improved} proposed a series of tricks namely feature matching, minibatch discrimination and historic averaging to stabilize the GAN training and avoid mode collapse. In WGAN~\cite{arjovsky2017wasserstein} and LSGAN~\cite{mao2017least} earth-mover and Pearson divergence distances are respectively used in the GAN objective instead of the JS-divergence, to induce stability in training which results in better mode discovery.  In~\cite{metz2016unrolled}, the generator objective is optimized by including the unrolled states of the discriminator, to address mode dropping. Drawing motivations from multi-agent learning, MADGAN \cite{ghosh2017multi} and GANMM \cite{yu2018mixture} employ the idea of using multiple generators and show that overall scheme performs similar to a mixture model with each generator learning one data mode. 

\section{Experiments and Results}

\subsection{Datasets and metrics}
We consider four image datasets namely, MNIST~\cite{lecun2010mnist}, FMNIST, CelebA~\cite{liu2015deep}, and CIFAR-10 \cite{krizhevsky2009learning} for experiments (qualitative illustration on a synthetic dataset is provided in the supplementary material, Fig. 1). Since the objective of all the  experiments is to obtain a well-clustered generation with class imbalance, we create imbalanced datasets from the standard datasets by either sub-sampling or merging multiple classes. Specifically, we consider the following data - (i) take two sets of two distinct MNIST classes, 0 Vs 4 (minimal overlap under t-SNE) and 3 Vs 5 (maximum overlap under t-SNE), with two different skews of 70:30 and 90:10, (ii) merge together `similar' clusters \{\{3,5,8\}, \{2\}, \{1,4,7,9\}, \{6\}, \{0\}\} to form a 5-class MNIST dataset (MNIST-5). Similarly, we also grouped FMNIST classes to create the FMNIST-5 dataset as \{\{Sandal, Sneaker, Ankle Boot\}, \{Bag\}, \{Tshirt/Top, Dress\}, \{Pullover, Coat, Shirt\}, \{Trouser\}\}, (iii) we consider CelebA dataset to distinguish celebrities with black hair from the rest. (iv) two classes of CIFAR (Frog Vs Planes, selected arbitrarily) with two synthetic imbalances of 70:30 and 90:10.  

We use Accuracy (ACC), Adjusted Rand Index (ARI) and Normalized Mutual Information (NMI)~\cite{fahad2014survey} as metrics to measure the clustering performance and Frechet Inception Distance (FID) \cite{miyato2018spectral} to measure the quality of the generated images. While the first three have to be higher, FID that quantifies the relative image quality of different models, have to be lower.

%Finally we conduct a curious experiment of discovering the unknown attributes of the data and share some interesting observations.

\subsection{Results and Discussions}

\begin{table}[]
\caption{Quantitative evaluation on imbalanced data for generation with clustering. Lower performances are observed with GANs where one of three conditions is violated.}
\centering
\begin{tabular}{lp{1.6cm}rrrr}
\hline
\textbf{Dataset} & \textbf{Model} & \textbf{ACC} & \textbf{NMI}       & \textbf{ARI}  & \textbf{FID}     \\ \hline
& $\hat{C}_1\hat{C}_2\hat{C}_3$ & 0.64      & 0.06  & 0.08     &    19.76 \\
&  $\mathbf{C_1}\hat{C}_2\hat{C}_3$    & 0.66 & 0.13 & 0.11 & 10.14   \\ 
MNIST-2 & $\hat{C}_1\mathbf{C_2}\hat{C}_3$
  & 0.75   & 0.20  & 0.25     &     15.11   \\ 
(70:30) 
& $\mathbf{C_1}\mathbf{C_2}\hat{C}_3$            & 0.81      & 0.40           & 0.38      &   4.63   \\ 
& $\mathbf{C_1}\mathbf{C_2}\mathbf{C_3}$  & \bf 0.98      & \bf 0.89 & \bf 0.93 & \bf 1.33\\ \hline

& $\hat{C}_1\hat{C}_2\hat{C}_3$          & 0.64      & 0.09            & 0.07     &     20.32 \\
&  $\mathbf{C_1}\hat{C}_2\hat{C}_3$   & 0.59 & 0.15 & 0.13 & 11.45   \\ 
MNIST-2 & $\hat{C}_1\mathbf{C_2}\hat{C}_3$
   & 0.61  & 0.24   & 0.25     &   10.84  \\ 
(90:10) 
& $\mathbf{C_1}\mathbf{C_2}\hat{C}_3$            & 0.77      & 0.33           & 0.54      &   6.08   \\ 
& $\mathbf{C_1}\mathbf{C_2}\mathbf{C_3}$  & \bf 0.98      & \bf 0.86 & \bf 0.91 & \bf 1.66\\ \hline

& $\hat{C}_1\hat{C}_2\hat{C}_3$          & 0.51 &0.21&0.19     &    20.64   \\
& $\mathbf{C_1}\hat{C}_2\hat{C}_3$   & 0.71 & 0.55 &  0.52 & 12.07         \\ 
MNIST-5 & $\hat{C}_1\mathbf{C_2}\hat{C}_3$
  & 0.76 & 0.59 &  0.64 & 15.31    \\ 
& $\mathbf{C_1}\mathbf{C_2}\hat{C}_3$            & 0.74      & 0.65           & 0.71      &   4.92   \\ 
& $\mathbf{C_1}\mathbf{C_2}\mathbf{C_3}$  & \bf 0.96 & \bf 0.89 & \bf 0.89 & \bf 1.13 \\ \hline

& $\hat{C}_1\hat{C}_2\hat{C}_3$  & 0.62 & 0.30 & 0.30    &     10.46  \\

& $\mathbf{C_1}\hat{C}_2\hat{C}_3$   & 0.77 & 0.66 & 0.61 & 5.41 
   \\ 
FMNIST-5 & $\hat{C}_1\mathbf{C_2}\hat{C}_3$
  & 0.75 & 0.68 & 0.65    &    9.20    \\ 

& $\mathbf{C_1}\mathbf{C_2}\hat{C}_3$            & 0.81      & 0.72           & 0.74      &   4.42   \\ 

& $\mathbf{C_1}\mathbf{C_2}\mathbf{C_3}$  & \bf 0.92 & \bf 0.81 & \bf 0.81 & \bf 0.69 \\ \hline
\end{tabular}
\label{tab:imbalance}
\end{table}

Results on MNIST-2 (3 Vs 5), MNIST-5 and FMNIST-5 are shown in Table~\ref{tab:imbalance}. It is observed that the GAN with all conditions satisfied (proposed) consistently outperforms the models that only satisfy the conditions partially, both in terms of cluster-purity and generation-quality. Similar observations are made on the colour datsets, CIFAR and CelebA as  summerized in Table~\ref{tab:imbalance1}. It is also seen that, the presence of the multimodal latent space ($C_1$) and a latent inverter ($C_2$) seem to affect the performance the most when there is class imbalance. This is corroborated by the fact that the performance of the  $\mathbf{C_1}\mathbf{C_2}\hat{C}_3$ model (ClusterGAN) is consistently best amongst all the models that partially satisfy the conditions. This implies that knowing the class-priors is an important pre-requisite to obtain a faithful clustered generation in GANs. 

Another important observation in CelebA experiment is that, different attributes e.g. eyeglasses, mustache etc., can divide the data into two clusters of different sizes. However, only black hair attribute divides the data into clusters of sizes 23.9\% and 76.1\% and by fixing the latent mode priors to 0.239 and 0.761, our model automatically discovers the black hair attribute and generates data accordingly. Finally, it is observed that the performance of the $\mathbf{C_1}\mathbf{C_2}\hat{C}_3$ and $\mathbf{C_1}\mathbf{C_2}\mathbf{C}_3$ are almost the same when the dataset is balanced (quantitative results in the supplementary material, Table IV). This is expected since the mode priors are matched by default in both the cases and the dataset has uniform priors. All these experiments suggests for a GAN model to generated well-clustered data, it should be equipped with all the stated conditions.
%Figure~\ref{sample5} and Figure~\ref{cifar2} show a few samples generated by our model for MNIST-5, FMNIST-5 and CIFAR-2 datasets. In Figure~\ref{sample5}, each column represents one cluster whereas in Figure~\ref{cifar2} each pane represent one cluster. It can been seen that clusters are pure in the sense that the images generated come only from the classes that are merged together. 
 %It is also observed in Table~\ref{tab:10class}, that the performance of the proposed method matches or better than the state-of-the-art methods on the entire datasets without class imbalance.
 
%Note that An interesting observation in CelebA experiment is that the latent space priors were fixed to 0.239 and 0.760
 
\begin{table}[]
\caption{Evaluation of the proposed method on colour datasets, CIFAR (Frogs Vs. Planes), CelebA (black hair Vs. non-black hair). It is seen that GANs that violate any of three required conditions offer lower performance.}
\centering
\begin{tabular}{lp{1.6cm}rrrr}
\hline
\textbf{Dataset} & \textbf{Model} & \textbf{ACC} & \textbf{NMI}       & \textbf{ARI}  & \textbf{FID}     \\ \hline

                 &   $\hat{C}_1\hat{C}_2\hat{C}_3$             & 0.66         & 0.41         & 0.46         & 55.54     \\
                 & $\mathbf{C_1}\hat{C}_2\hat{C}_3$            & 0.70         & 0.51         & 0.54         & 42.87        \\
CIFAR-2          &$\hat{C}_1\mathbf{C_2}\hat{C}_3$
           & 0.75         & 0.60         & 0.63         & 47.35          \\
(70:30)         
                & $\mathbf{C_1}\mathbf{C_2}\hat{C}_3$            & 0.72      & 0.68           & 0.65      &   44.38   \\ 
                 & $\mathbf{C_1}\mathbf{C_2}\mathbf{C_3}$ & \bf 0.88     & \bf 0.70     & \bf 0.75     & \bf 31.15    \\ \hline
                 &   $\hat{C}_1\hat{C}_2\hat{C}_3$             & 0.50         & 0.19         & 0.17         & 58.45     \\
                 & $\mathbf{C_1}\hat{C}_2\hat{C}_3$            & 0.54         & 0.18         & 0.29         & 43.87       \\
CIFAR-2          & $\hat{C}_1\mathbf{C_2}\hat{C}_3$
           & 0.63         & 0.26         & 0.39         & 48.16          \\
(90:10)         
                & $\mathbf{C_1}\mathbf{C_2}\hat{C}_3$            & 0.67      & 0.22           & 0.21      &   42.01   \\ 
                 & $\mathbf{C_1}\mathbf{C_2}\mathbf{C_3}$ & \bf 0.83     & \bf 0.26       & \bf 0.28       & \bf 32.86    \\ \hline
& $\hat{C}_1\hat{C}_2\hat{C}_3$      & 0.55 & 0.02 & 0.01 & 150.2  \\
& $\mathbf{C_1}\hat{C}_2\hat{C}_3$   & 0.58 & 0.15 & 0.14 & 110.9 \\ 
CelebA & $\hat{C}_1\mathbf{C_2}\hat{C}_3$
  & 0.57 & 0.14 & 0.23   & 83.56  \\ 

& $\mathbf{C_1}\mathbf{C_2}\hat{C}_3$            & 0.64      & 0.18           & 0.26      &  67.1    \\ 
& $\mathbf{C_1}\mathbf{C_2}\mathbf{C_3}$ & \bf 0.81 & \bf 0.30 & \bf 0.38 & \bf  62.9 \\ \hline
\end{tabular}

\label{tab:imbalance1}
\vspace{-0.4cm}
\end{table}

\subsection{Results for Prior learning}
As mentioned in Section III.A, the proposed method of latent construction with latent inverter could be used to learn the class-priors (if unknown) with sparse supervision (note that the clustering experiments are completely independent of prior-learning where the priors were assumed to be either uniform or known a-priori).
To evaluate the performance of the proposed prior learning method, we consider the same setup as in the previous section, with imbalanced class priors. We initialize $\boldsymbol{\alpha}$ uniformly with same value for each of its element. Priors are learned with the technique described in Section III.B.2 using 1\% of the labelled data.  The learned priors are compared with real data priors in Table~\ref{tab:prior}. It is seen that  the proposed technique learns class priors accurately for all the cases considered. 

% \begin{figure}[!t]
% \centering
% \includegraphics[width=2.5cm]{mnist5p.png}
% \hspace{0.5cm}
% \includegraphics[width=2.5cm]{fmnist5p.png}
% \caption{Samples generated by our model for MNIST-5 (left) and FMNIST-5 (right) data. Each column represents one mode. The generated modes clearly capture the grouping of the classes.}
% \label{sample5}
% \end{figure}

% \begin{figure}[!t]
% \centering
% \includegraphics[width=3.5cm]{class_0_02.png}
% \hspace{0.5cm}
% \includegraphics[width=3.5cm]{class_1_04.png}
% \caption{Samples generated by the proposed method for CIFAR-2, frogs and planes, (70:30) data. Each pane represents one mode.}
% \label{cifar2}
% \end{figure}

%we consider the variation in imbalance ratio of digits 4 and 0 from 10:90 to 50:50 and observe the prior learned by the proposed method. Observations are plotted in Figure~\ref{p_learn}, learned priors can be seen matching closely with the real priors. Real priors are nothing but the class imbalance ratios. Note that in each experiment, initial value of the priors are set be uniform or equal for both the classes. The model learns the desired prior values during the training. Considering all the aforementioned results, the proposed method seems significant in that it can cluster the generated data in GANs with only a less than a percent of labelled data. 

\begin{table}[H]
\caption{Evaluation of the proposed prior learning method.}
\centering

\begin{tabular}{lcc}
\hline
\textbf{Dataset} & \textbf{Real data priors} & \textbf{Learned priors}     \\ \hline
MNIST-2 & [0.7, 0.3]     & [0.709, 0.291]           \\ 
MNIST-2 & [0.9, 0.1]     & [0.891, 0.109]           \\ 
MNIST-5 & [0.3, 0.1, 0.4,      & [0.291, 0.095, 0.419,         \\ 
& 0.1, 0.1] & 0.095, 0.099]   \\
FMNIST-5 & [0.3, 0.1, 0.3,      & [0.304, 0.096, 0.284,         \\ 
& 0.2, 0.1] & 0.220, 0.096]   \\
CIFAR-2 & [0.7, 0.3]     & [0.679, 0.321]           \\ 
CIFAR-2 & [0.9, 0.1]     & [0.876, 0.124]           \\  
CelebA-2 & [0.239, 0.761]     & [0.272, 0.727]           \\ \hline
\end{tabular}

\label{tab:prior}
%\vspace{-0.4cm}
\end{table}

\section{Conclusion}
In this work, we described the problem of clustering in the generated space of GANs and investigated the role of latent space characteristics in obtaining the desired clustering. We showed, this can be achieved by having a multimodal latent space along with a latent space inversion network and matched priors of latent and real data distribution. We also proposed to parameterize the latent space such that its characteristics can be learned. It also leads to the development of a technique for learning the unknown real data class-priors using sparse supervision. 
%of in a way that the class priors can be learned and imposed on the latent space, using minimal-supervision ($<1\%$), which enables matching of the statistical properties of latent and data distributions. 
Our analysis results in a GAN model which offers the advantages of robust generation under the setting of skewed data distributions and clustering, where the existing methods showed sub-optimal performances. To the best of our knowledge, this is the first work that demonstrates the importance of latent structure on the ability of GANs to generate well-clustered data. %In future, we will explore the possibility of applying our method to non-image datasets.

{\small
\bibliographystyle{ieee_fullname}
\bibliography{egbib}
}

\begin{figure*}[!t]
\centering
\begin{minipage}[]{0.16\linewidth}
  \centerline{\includegraphics[width=3.15cm]{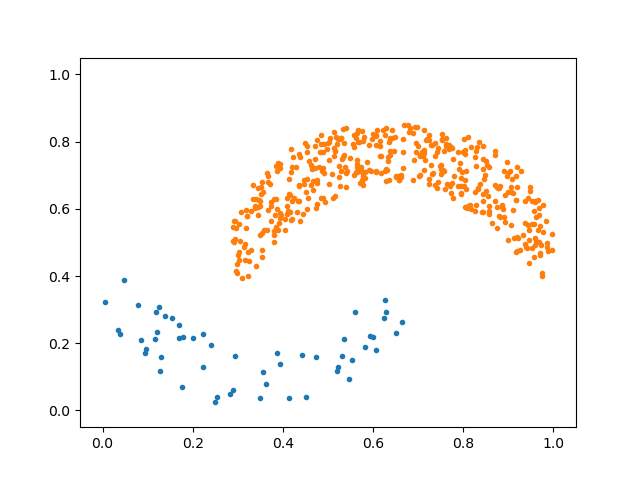}}
  \centerline{\footnotesize{(a) Real data} }
\end{minipage}
\begin{minipage}[]{.16\linewidth}
  \centerline{\includegraphics[width=3.15cm]{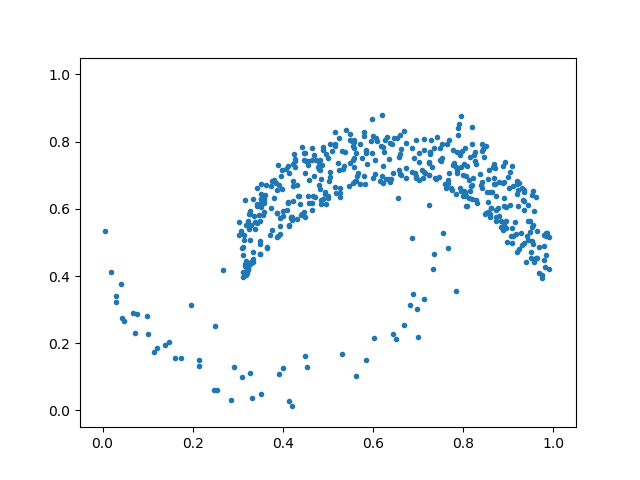}}
  \centerline{\footnotesize(b) WGAN}%\medskip
\end{minipage}
\begin{minipage}[]{0.16\linewidth}
  \centerline{\includegraphics[width=3.15cm]{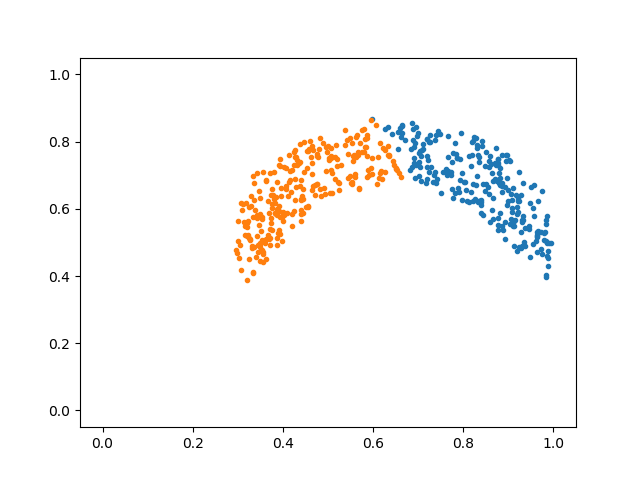}}
  \centerline{\footnotesize{(c) DeliGAN} }
\end{minipage}
\begin{minipage}[]{.16\linewidth}
  \centerline{\includegraphics[width=3.15cm]{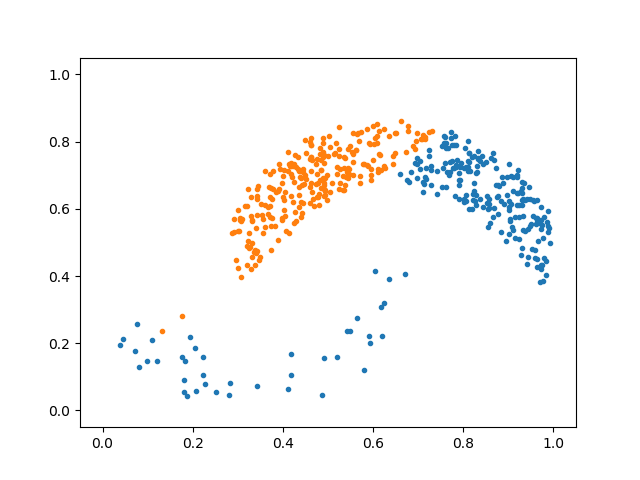}}
  \centerline{\footnotesize(d) InfoGAN}%\medskip
\end{minipage}
\begin{minipage}[]{0.16\linewidth}
  \centerline{\includegraphics[width=3.15cm]{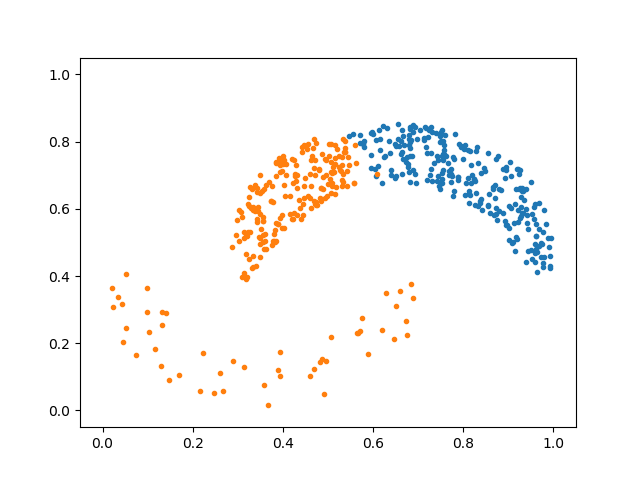}}
  \centerline{\footnotesize{(e) ClusterGAN} }
\end{minipage}
\begin{minipage}[]{.16\linewidth}
  \centerline{\includegraphics[width=3.15cm]{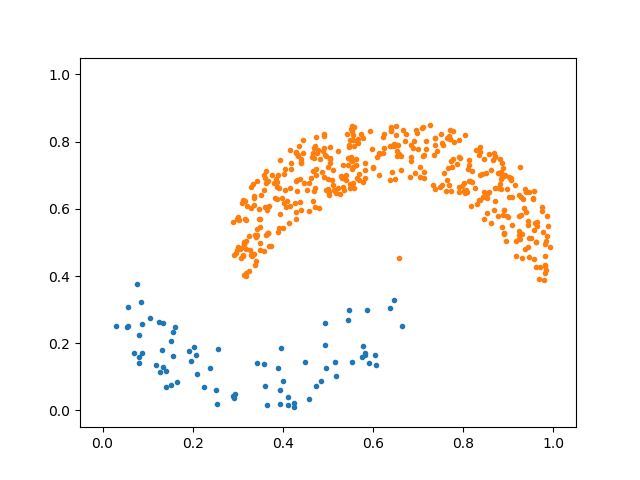}}
  \centerline{\footnotesize(f) Ours}%\medskip
\end{minipage}
\caption{Clustering in the generated spaces produced by different GANs for two-class Moon-data with cluster size ratio of 80:20. In absence of multimodal latent space, latent inverter and prior matching (WGAN), entire data is confined to a single cluster (Impossible conditional generation). Fulfilment of only one of the three requirements, e.g. only multimodal latent space (DeliGAN) can generate two classes but misses one of the clusters completely. Similarly the presence of multimodal latent space and latent space inverter (InfoGAN and ClusterGAN) are also unable to provide desired clustering in absence of matched priors. Our method satisfies all three conditions and thus can faithfully cluster.}
\label{moon}
\vspace{-0.3cm}
\end{figure*}

% \begin{figure*}[h]
% \centering
% {\includegraphics[width=6 in]{Moon.png}}
% \end{figure*}

\newpage

\begin{figure}[!t]
\centering
{\includegraphics[width=8cm]{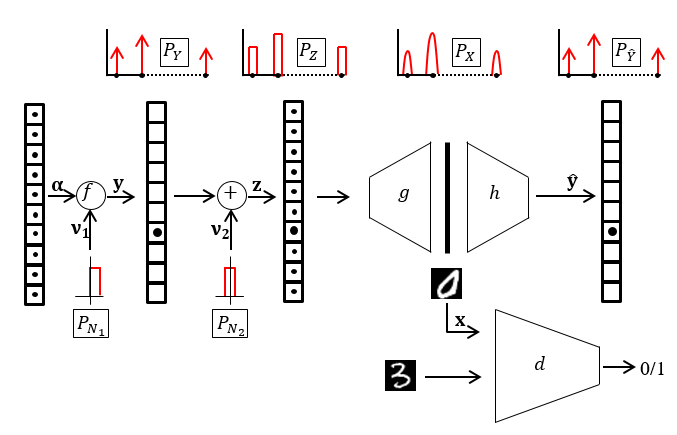}}
\caption{Illustration of the proposed pipeline for clustering. Generator $g(\rvz)$ tries to mimic the real data distribution $P_{X}$ with the help of discriminator $d$. The inversion network $h(g(\rvz))$ inverts the generation process to ensure the matching of clustering properties of generating and latent distributions. Mode priors of the latent space is encoded in $\rvy$ by reparameterizing a known distribution $P_{N_1}({\nu_1})$ using a learnable vector $\boldsymbol{\alpha}$.}
\label{bd}
\end{figure}

\section{Additional Proofs}

\textbf{Proof for Corollary 1.2:} Since both $g$ and $h$ are continuous mappings (neural networks) and supports of all the distributions are disjoint,
\begin{equation}
    \int_{{\sZ}_i}P_Z~d\rvz = \int_{{\sX}_i}P_X~d\rvx=\int_{{\hat{\sY}}_i}P_{\hat{Y}}~d\mathbf{\hat{y}}
\end{equation} and, 
\begin{equation}
    \int_{{\sW}_i}P_W~d\rvw \neq \int_{{\sZ}_i}P_Z~d\rvz \implies \int_{{\sW}_i}P_W~d\rvw \neq \int_{{\sX}_i}P_X~d\rvx
\end{equation}

\textbf{Proof for Lemma 2:} Since $\sigma_h$ is a unit step function, $f$ is the first order difference or discrete Dirac delta function positioned at $a_i$.
%\begin{equation}
%    f_{1_i} = \delta(a_i - \nu_1)
%    \label{delta}
%\end{equation}
Now by definition,
\begin{equation}
    P_Y(\rvy=i) = P(f_{i} \neq 0)
    \label{prior}
\end{equation}
From \eqref{f1}, we can see that $f_{i}$ becomes non-zero only for $a_{i-1}\leq\nu_1\leq a_i$, therefore,

\begin{align}
P_Y(\rvy=i) &= P_{N_1}(a_{i-1}\leq\nu_1\leq a_i)\\
     &= \int_{a_{i-1}}^{a_i}P_{N_1}(\nu)d\nu = a_i - a_{i-1} = \frac{e^{\alpha_i}}{\sum_{k}{e^{\alpha_k}}}\label{alpha}
   \end{align}
   
\textbf{Proof for Lemma 3:} 
\begin{align}
\KL (P_{\hat{Y}} || P_{Y}) &= \sum_{\hat{\rvy}=i} {P_{\hat{Y}}}\log \frac{P_{\hat{Y}}}{P_Y} \quad\quad s.t. \quad i\in \{0, 1\} 
\label{kl1}\\
&= \sum_{\hat{\rvy}=i} \left({P_{\hat{Y}}}\log {P_{\hat{Y}}} - {P_{\hat{Y}}}\log {P_Y}\right)
\label{kl2}\\
%&= \sum_{\hat{\rvy}} {P_{\hat{Y}}}\log {P_{\hat{Y}}} - {P_{\hat{Y}}({\hat{\rvy}} = 0 )}\log {P_Y(\rvy=0)} - {P_{\hat{Y}}({\hat{\rvy}} = 1 )}\log {P_Y(\rvy=1)}
%\label{kl3}\\
&= \sum_{\hat{\rvy}=i} \left({P_{\hat{Y}}}\log {P_{\hat{Y}}} - {P_{\hat{Y}}}\log {\int_{\sZ_i}P_{Z}~d\rvz} \right)
\label{kl4}
\end{align}
Since $\int_{ \sZ_i }P_{Z} ~d\rvz = \int_{\sX_i}P_X~d\rvx$, \eqref{kl2} can be written as
\begin{equation}
\KL (P_{\hat{Y}} || P_{Y}) = \sum_{\hat{\rvy}=i} \left({P_{\hat{Y}}}\log {P_{\hat{Y}}} - {P_{\hat{Y}}}\log {\int_{\sX_i}P_X~d\rvx} \right)
\label{kl5}
\end{equation}
Since  $\int_{{\sX}_i}P_X~d\rvx = P_{\hat{X}}(\hat{\rvx} =i )$, by definition, \eqref{kl5} can be written as
\begin{align}
%\KL (P_{\hat{Y}} || P_{Y}) &= \sum_{\hat{\rvy}} {P_{\hat{Y}}}\log {P_{\hat{Y}}} - {P_{\hat{Y}}({\hat{\rvy}} = 0 )}\log {P_{\hat{X}}(\hat{\rvx} =0 )} - {P_{\hat{Y}}({\hat{\rvy}} = 1 )}\log{P_{\hat{X}}(\hat{\rvx} =1)}
%\label{kl6}\\
\KL (P_{\hat{Y}} || P_{Y}) &= \sum_{\hat{\rvy}=i} \left({P_{\hat{Y}}}\log {P_{\hat{Y}}} - {P_{\hat{Y}}}\log {P_{\hat{X}}}\right)
\label{kl6.1}\\
&= \KL (P_{\hat{Y}} || P_{\hat{X}})
\label{kl7} 
\end{align}

\begin{table}[h]
\caption{Quantitative evaluation on balanced data for generation with clustering. Multimodal latent space with latent inverter offers similar performance as model with all three conditions satisfied when the data is balanced. %It can be seen that proposed scheme outperforms the benchmarks even in the case of balanced classes.
}
\centering
\begin{tabular}{llrrrr}
\hline
\textbf{Dataset} & \textbf{Model} & \textbf{ACC} & \textbf{NMI}       & \textbf{ARI}  & \textbf{FID}     \\ \hline
& $\hat{C}_1\hat{C}_2\hat{C}_3$         & 0.64      & 0.61            & 0.49     &     10.83 \\
& $\mathbf{C}_1\hat{C_2}\hat{C}_3$
      & 0.89         & 0.86               & 0.82     &   8.74   \\ 
MNIST & $\hat{C_1}\mathbf{C}_2\hat{C}_3$   &0.89&0.90&0.84& 7.34 \\
&  $\mathbf{C_1}\mathbf{C_2}\hat{C}_3$ & 0.95         & 0.89               & 0.89     &   1.84       \\ 
& $\mathbf{C_1}\mathbf{C_2}\mathbf{C_3}$  & \bf 0.96      & \bf 0.91 & \bf 0.92 & \bf 1.82\\ 

\hline
& $\hat{C}_1\hat{C}_2\hat{C}_3$   & 0.34      & 0.27            & 0.20        & 19.80\\
& $\mathbf{C_1}\hat{C}_2\hat{C}_3$        & 0.61         & 0.59                & 0.44& 12.44          \\ 
FMNIST &  $\hat{C_1}\mathbf{C}_2\hat{C}_3$   & 0.55 & 0.60 & 0.44 & 6.95     \\
%& DEC           & 0.13      & 0.11           & 0.11            \\ 
& $\mathbf{C_1}\mathbf{C_2}\hat{C}_3$    & 0.63         & 0.64               & 0.50    & 0.56              \\
& $\mathbf{C_1}\mathbf{C_2}\mathbf{C_3}$        & \bf 0.65      & \bf 0.70 & \bf 0.63 & \bf 0.55\\

\hline
& $\hat{C}_1\hat{C}_2\hat{C}_3$   & 0.24      & 0.36            & 0.26        & 46.80\\
& $\mathbf{C_1}\hat{C}_2\hat{C}_3$        & 0.43         & 0.39                & 0.46 & 40.44          \\ 
CIFAR &  $\hat{C_1}\mathbf{C}_2\hat{C}_3$   & 0.52 & 0.42 & 0.48 & 36.95     \\
%& DEC           & 0.13      & 0.11           & 0.11            \\ 
& $\mathbf{C_1}\mathbf{C_2}\hat{C}_3$    & 0.60         & 0.68               & 0.69    & 29.66              \\
& $\mathbf{C_1}\mathbf{C_2}\mathbf{C_3}$        & \bf 0.67      & \bf 0.76 & \bf 0.72 & \bf 26.35\\\hline
\end{tabular}

\label{tab:10class}
\vspace{-0.4cm}
\end{table}

\section{Additional Experiments}

\subsection{D. Mode Separation}

\begin{figure}[h]
\centering
\begin{minipage}{0.4\linewidth}
  \centerline{\includegraphics[width=4cm]{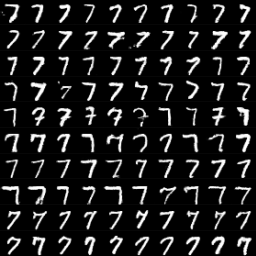}}
  %\centerline{\footnotesize{(a) Real images from MNIST} }
\end{minipage}
\begin{minipage}{0.4\linewidth}
  \centerline{\includegraphics[width=4cm]{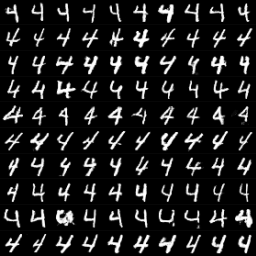}}
  %\centerline{\footnotesize{(b) Images inferred using BiGAN} }
\end{minipage}
\caption{Demonstration of mode separation using the proposed method. Every row in each figure depicts sample from a mode when the the proposed method is trained only with a single digit type with a latent space with ten modes.}
\label{semantics}
\end{figure}

In this work, semantics of the data refer to the modes in data distribution. These semantics represent different attributes of the samples and are separated out by the proposed method. For a better understanding, experiments are conducted with samples of only a single digit type from the MNIST dataset. Samples of digit 7 and 4 are considered for this purpose. The proposed GAN ($\mathbf{C_1}\mathbf{C_2}\mathbf{C}_3$) is trained with a discrete uniform latent space with 10 modes and the generated images are shown in Fig.~\ref{semantics}. Each row in Fig.~\ref{semantics} corresponds on one latent space mode and shows different attributes of the considered digits. For example, the fifth row in left pane contains generated images of digit 7 with slits. Similarly in right pane, the third row contains images of digit 4 with a closed notch. Note that, even with images of a single digit, no mode collapse is observed with the proposed method.

\subsection{E. Attribute discovery}
\label{a_disc}

\begin{figure}[h]
\centering
\begin{minipage}[]{0.4\linewidth}
  \centerline{\includegraphics[width=4.2cm]{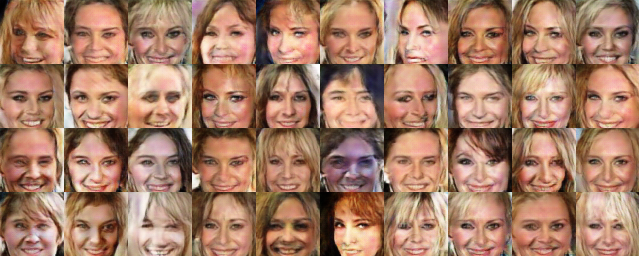}}
  %\centerline{\footnotesize{(a) 75:25} }
\end{minipage}
\begin{minipage}[]{0.4\linewidth}
  \centerline{\includegraphics[width=4.2cm]{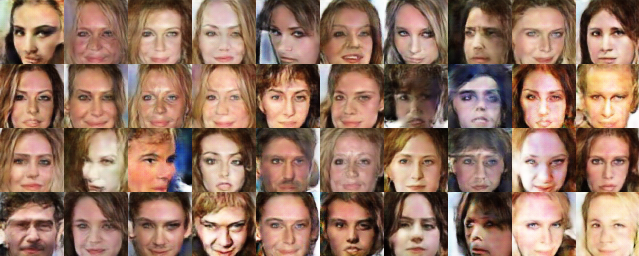}}
  %\centerline{\footnotesize{(b) 75:25} }
\end{minipage}
\caption{Discovery of the facial attribute smile with teeth visible. Sample images generated in the experiments with class ratio of 70:30 for faces from the CelebA dataset.}
\label{gen_cl_t}
\end{figure}

\begin{figure}[h!]
\centering
\begin{minipage}[]{0.3\linewidth}
  \centerline{\includegraphics[width=3.2cm]{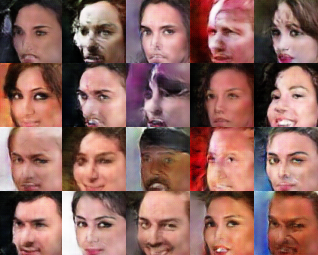}}
  %\centerline{\footnotesize{(a) 50:50} }
\end{minipage}
\begin{minipage}[]{0.3\linewidth}
  \centerline{\includegraphics[width=3.2cm]{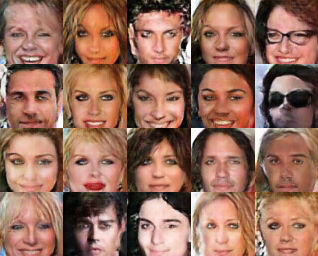}}
  %\centerline{\footnotesize{(b) 25:75} }
\end{minipage}
\begin{minipage}[]{.3\linewidth}
  \centerline{\includegraphics[width=3.2cm]{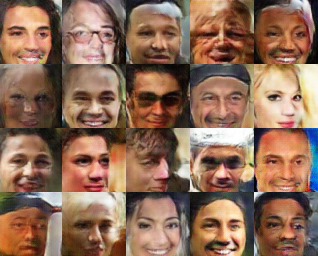}}
  %\centerline{\footnotesize(c) 02:98}%\medskip
\end{minipage}
\caption{Discovery of the attribute facial pose-angle. Sample images generated in the experiments with class ratio of 10:30:60 for from the CelebA dataset.}
\label{gen_cl_s}
\end{figure}

 In a few real-life scenarios, the class imbalance ratio is unknown. In such cases, an unsupervised technique should discover semantically plausible regions in the data space. To evaluate the proposed method's ability to perform such a task, we perform experiments where sample from $P_Y$ are drawn with an assumed class ratio rather than a known ratio.
 Two experiments are performed on CelebA, first with the assumption of 2 classes having a ratio of 70:30 and the second with the assumption of 3 classes having a ratio of 10:30:60. In the first experiment, the network discovers visibility of teeth as an attribute to the faces whereas in the second it learns to differentiate between the facial pose angles. Conditional generation from both the experiments are shown in figure \ref{gen_cl_t} and \ref{gen_cl_s}, respectively. Note that these attributes are not labelled in the dataset but are discovered by our model.

\subsection{F. Mode counting using proposed method}

\begin{figure}[h]
\centering
\begin{minipage}{0.4\linewidth}
  \centerline{\includegraphics[width=5cm]{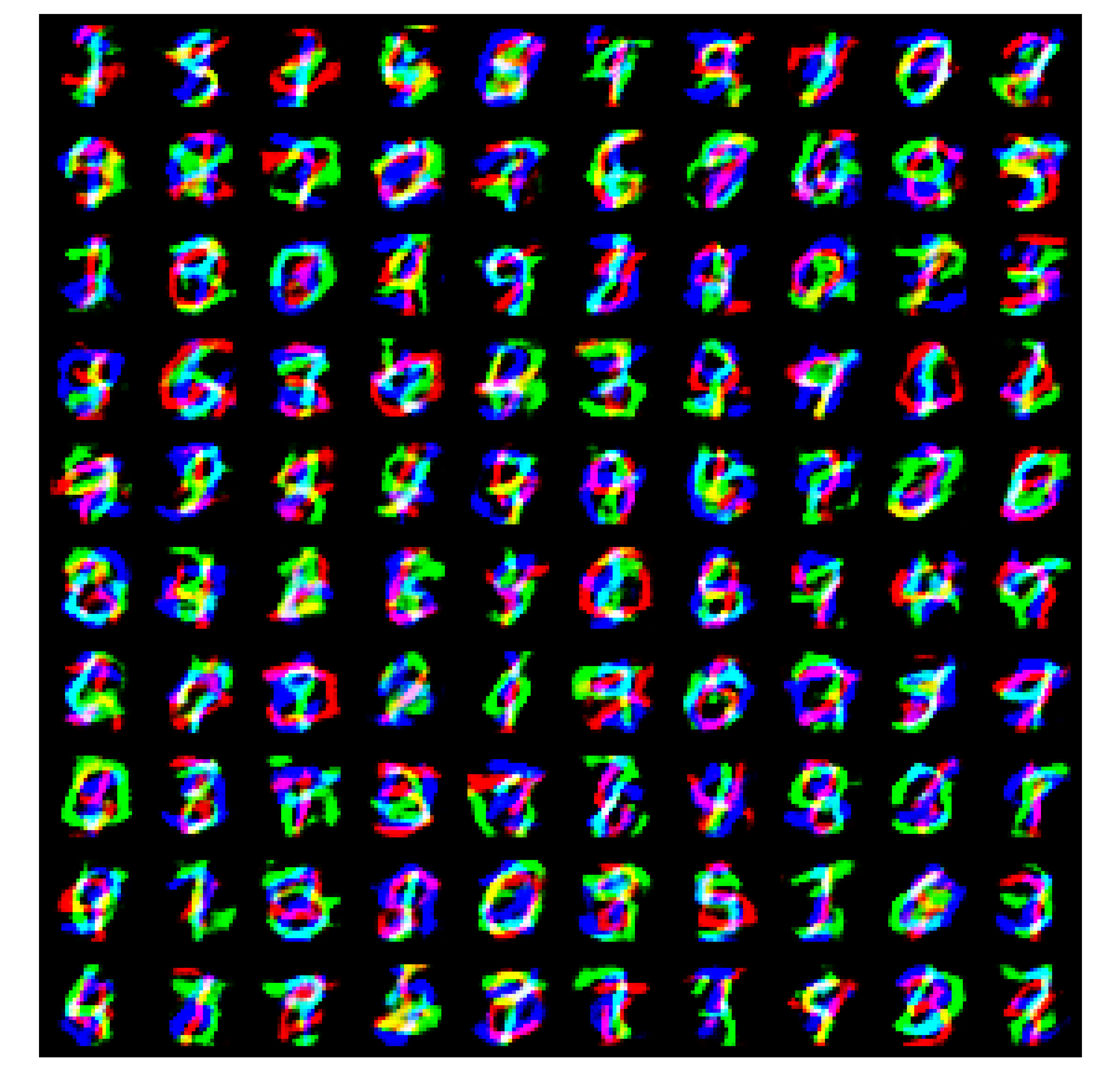}}
  %\centerline{\footnotesize{(a) Real images from MNIST} }
\end{minipage}
\caption{Mode counting experiment result for stacked MNIST dataset. The proposed method is able to produce variety of modes after training.}
\label{stack}
\end{figure}

\begin{figure}[h]
\centering
\begin{minipage}{0.4\linewidth}
  \centerline{\includegraphics[width=3.5cm]{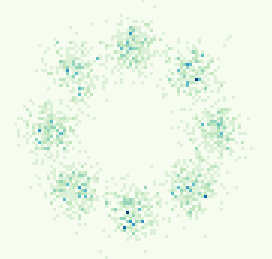}}
  \centerline{\footnotesize{(a) Real data} }
\end{minipage}
\begin{minipage}{0.4\linewidth}
  \centerline{\includegraphics[width=3.5cm]{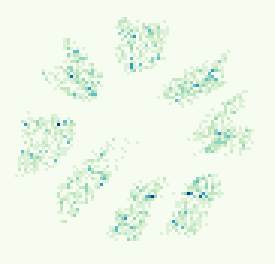}}
  \centerline{\footnotesize{(a) Generated data} }
\end{minipage}
\caption{Density plots of true data and the proposed method's generator output for 8 component GMM arranged over a circle}
\label{ring}
\end{figure}

We trained the proposed method for mode counting experiment on stacked MNIST dataset. It is able to generate 993 modes. Some of the generated images are shown in Fig.~\ref{stack}. Similar performance is observed in 8 component GMM experiment, as shown in Fig.~\ref{ring}.

% \begin{figure}[t!]
% \centering
% \begin{minipage}{0.4\linewidth}
%   \centerline{\includegraphics[width=3.5cm]{step_hmoon.png}}
%   \centerline{\footnotesize{(a) Real data} }
% \end{minipage}
% \begin{minipage}{0.4\linewidth}
%   \centerline{\includegraphics[width=3.5cm]{step_gen_hmoon.png}}
%   \centerline{\footnotesize{(a) Generated data} }
% \end{minipage}
% \caption{Density plots of true data and the proposed method generator output for two classes arranged in an interleaved crescent pattern. }
% \label{ring}
% \end{figure}

%

\end{document}

%% file: math_commands.tex
\usepackage{amsmath,amsfonts,bm}

\def\eqref#1{equation~\ref{#1}}
\def\1{\bm{1}}

\def\rvw{{\mathbf{w}}}
\def\rvx{{\mathbf{x}}}
\def\rvy{{\mathbf{y}}}
\def\rvz{{\mathbf{z}}}

\DeclareMathAlphabet{\mathsfit}{\encodingdefault}{\sfdefault}{m}{sl}
\SetMathAlphabet{\mathsfit}{bold}{\encodingdefault}{\sfdefault}{bx}{n}

\def\sR{{\mathbb{R}}}

\def\sW{{\mathbb{W}}}
\def\sX{{\mathbb{X}}}
\def\sY{{\mathbb{Y}}}
\def\sZ{{\mathbb{Z}}}

\newcommand{\E}{\mathbb{E}}

\newcommand{\KL}{D_{\mathrm{KL}}}

\DeclareMathOperator*{\argmax}{arg\,max}